\documentclass{article}
\usepackage{spconf,amsmath,graphicx}
\usepackage{times}
\usepackage{url}
\usepackage{multirow}
\usepackage{tipa}
\usepackage{booktabs,caption,threeparttable}
\usepackage{xcolor}
\usepackage{todonotes}
\usepackage{amsmath}
\usepackage{amssymb}
\usepackage[hidelinks]{hyperref}
\usepackage{float}

\title{Combining Contrastive and Non-Contrastive Losses \\
for Fine-Tuning Pretrained Models in Speech Analysis}
\name{Florian Lux, Ching-Yi Chen, Ngoc Thang Vu}
\address{University of Stuttgart, Institute for Natural Language Processing, Germany}
\copyrightnotice{978-1-6654-7189-3/22/\$31.00~\copyright2023 IEEE}

\begin{document}
%
\maketitle
\begin{abstract}
Embedding paralinguistic properties is a challenging task as there are only a few hours of training data available for domains such as emotional speech. One solution to this problem is to pretrain a general self-supervised speech representation model on large amounts of unlabeled speech. This pretrained model is then finetuned to a specific task. Paralinguistic properties however have notoriously high class variance, making the finetuning ineffective. In this work, we propose a two step approach to this. First we improve the embedding space, then we train an adapter to bridge the gap from the embedding space to a classification task. In order to improve the class invariance we use a combination of contrastive and non-contrastive losses to explicitly optimize for class invariant, yet discriminative features. Our approach consistently outperforms baselines that are finetuned end-to-end on multiple tasks and surpasses a benchmark on state-of-the-art emotion classification.

\end{abstract}
\noindent\textbf{Index Terms}: speech embedding, contrastive, non-contrastive, representation adjustment

\section{Introduction and Related Work}
Extracting states and traits of speakers based on their speech is a necessity for many applications. Examples of this include identifying a speaker by their voice to personalize spoken human computer interfaces with user profiles \cite{lu2011speakersense} or for forensics \cite{rose2002forensic}, verifying the identity of a speaker through their voice as a form of biometric password \cite{singh2018voice}, and even clinical applications such as the detection of respiratory illnesses \cite{cummins2017you, shrivastava2018assessment} and even less saliently linked properties, such as the detection of depression from speech \cite{nasir2016multimodal}. But not only the classification of such properties is an immensely important topic, the latent space of models performing such tasks is also an invaluable tool for transfer learning certain properties in other tasks. For example, the ability to synthesize speech from even unseen speakers in text-to-speech is usually transfer learned from the speaker identification task \cite{jia2018transfer}. Similarly, the cocktail party problem in automatic speech recognition can be solved very well through transfer learning from speaker identification \cite{denisov2019end}.

Supervised methods that are highly specialized for these tasks generally exhibit the best performances, for example the recent ECAPA-TDNN architecture \cite{ecapa}, or x-Vector \cite{xvect} and d-Vector \cite{dvector} embeddings. However, many domains,  such as speech emotion recognition (SER), struggle with underlying fuzzy concepts and a lack of large training datasets. To remedy this, self-supervised learning of general purpose speech representations is frequently applied \cite{Hsu2021HuBERTSS, ling2020decoar, ling2020deep, liu2020mockingjay, liu2021tera}. Two remarkable self-supervised frameworks are wav2vec 2.0 \cite{baevski2020wav2vec} and HuBERT \cite{Hsu2021HuBERTSS}. They are both competitive methods for learning latent representations in a language-model-like fashion. While they share one architecture, the way they are trained differs in that HuBERT employs mechanisms to better handle the nature of sequential units in speech. The general representations that such a pretrained model produces however still need task-specific finetuning to achieve the best performance in a specific downstream task, such as speaker verification \cite{Fan2021ExploringW2}, speech emotion classification \cite{pepino21_interspeech}, or automatic speech recognition \cite{yi2020applying}. 

Another approach that can work self-supervised is the contrastive learning framework, that aims to group positive (i.e. similar) samples closer and distance negative (i.e. different) samples further from each other in the latent space \cite{bromley1993signature, oord2019representation, 10.1007/978-3-319-24261-3_7}. Recently the Barlow Twins approach \cite{zbontar2021barlow} has been proposed, which is non-contrastive and works solely by explicitly reducing class variance through its objective function. It is competitive with state-of-the-art methods for self-supervised learning in the image domain, but has not been applied to the speech domain as of yet. A common problem that the model family of contrastive and non-contrastive learning is very prone to suffer from is mode-collapse \cite{dieng2019avoiding}, i.e., finding an unsatisfactory trivial solution by having all datapoints be very similar to each other. This can be fixed by applying additional methods and constraints \cite{Chong2020SimpleAE, 9312049}. The Barlow Twins approach however solves mode-collapse via its redundancy reduction mechanism, which ensures that every dimension in the latent space is informative and non-overlapping with other dimensions.

In this work, we combine a self-supervised model for general speech representation with contrastive and non-contrastive methods for finetuning to different downstream tasks which leads to improved performance over naive end-to-end finetuning. We use the triplet \cite{10.1007/978-3-319-24261-3_7} and Barlow Twins \cite{zbontar2021barlow} losses in combination to explicitly improve distinctiveness and class invariance in the embedding space of fuzzy tasks such as SER or speaker identification (SID) with varied recording setups and degrees of expressiveness within speaker classes.

Furthermore, we propose a two-stage finetuning framework. In the first step, we finetune the speech representations for a certain task using contrastive and non-contrastive learning methods and in a second step we adapt the tuned speech embeddings fully to the task with a simple classifier.

We validate the quality of generated embeddings in a wide range of downstream speech analysis tasks such as SER and SID, as well as age and gender classification by measuring task performance. On SER, we outperform all wav2vec 2.0 based models in the SUPERB benchmark \cite{yang21c_interspeech} using the wav2vec 2.0 base model as the embedding backbone. We also use the HuBERT base model as the backbone and outperform the corresponding model in the SUPERB benchmark, stay however slightly below the large variant of the HuBERT model while using only the base variant as the backbone. In all of those cases we do so without performing hyperparameter searches using the respective backbone, as would be necessary to get the best performance in a challenge setting. Furthermore we evaluate cluster integrity which is directly linked to linear classification performance, but is also important for zero-shot applications under the learn-to-compare framework \cite{sung2018learning}, that treats classification as clustering in latent space, such as k-nearest-neighbors \cite{friedman1975algorithm} in latent space (i.e. Siamese Networks \cite{bromley1993signature}) or k-means \cite{lloyd1957least} in latent space (i.e. Prototypical Networks \cite{snell2017prototypical}).  

Our contributions can be summarized as follows: 1) We propose a novel two-step finetuning framework that combines contrastive and non-contrastive losses to improve intermediate representations before feeding them into a classifier and demonstrate its success in a wide range of speech analysis tasks. 2) We explore the use of the Barlow Twins approach \cite{zbontar2021barlow} in the speech domain for the first time and introduce modifications that make it work on speech analysis tasks. 
For a convenient lookup of our exact setup as well as to ensure reproducibility, we release all of our code open source\footnote{\url{https://github.com/DigitalPhonetics/BetterFinetuning}}.

\section{Method}
This section illustrates the three proposed approaches in two aspects: The system architecture and the loss functions. All systems consist of a finetuned encoder and a task-dependent adapter as shown in Figures \ref{fig:contrasitive} and \ref{fig:BarlowTwins}. First we finetune the encoder that has been pretrained to produce general purpose speech representations to generate embeddings that are adjusted to the domain. In the second phase, we freeze the encoder and use a classifier/adapter to leverage the speech embedding in downstream tasks. Details of the downstream tasks will be presented in section \ref{sec:experiments}, but the method itself is task-agnostic and theoretically not even limited to the speech modality.

\subsection{System Description}
We use a publicly available pretrained wav2vec 2.0 model\footnote{\url{https://huggingface.co/facebook/wav2vec2-base}} \cite{baevski2020wav2vec} as well as a publicly available HuBERT model\footnote{\url{https://huggingface.co/facebook/hubert-base-ls960}} \cite{Hsu2021HuBERTSS} as our pre-trained encoder. Only one of the two encoders is used at a time, but to show that our approach is mostly agnostic to the pretrained general model, we experiment using either one as the backbone. Audio inputs of arbitrary length are fed into the encoder and then averaged over the time-axis to reduce them to a single vector. The sequence nature of the data is however not lost through this, since both wav2vec 2.0 as well as HuBERT learn to encode speech representations in a highly contextualized manner. Both the wav2vec 2.0 model and the HuBERT model do so by randomly masking feature vectors before passing them to a transformer network. The difference mostly lies in the fact that wav2vec 2.0 learns its targets simultaneously while training the model, whereas HuBERT builds its targets via a separate clustering process, that is subsequently used to apply the prediction loss only over the masked sequences. The authors state, that this forces the model to learn a combined acoustic and language model to deal with the segmental nature of units in speech. So in theory, HuBERT should be a lot better suited for tasks that rely on recognizing segmental units in speech (e.g. phones, intonation phrases, intermediate phrases etc.). While this is not necessarily the case for many speech analysis tasks, where the prosody is often more important than the segmental information, the HuBERT model does show consistent improvements over wav2vec 2.0 in multiple speech analysis tasks in the SUPERB benchmark \cite{yang21c_interspeech}.
Regardless of the backbone used, the averaged encoded sequence is then projected into a smaller dimensionality to help enforce generalization on relatively small amounts of data (i.e. often less than 10 hours) through a bottleneck, which concludes the first part of the architecture. 

In the second part, the embeddings are further fed to an adapter network which consists of 2 fully connected layers with ReLU activations. All approaches share the same architecture of first the backbone, followed by the adapter. The difference between them lies in how we finetune the pretrained encoder from its general representations to ones that suit our target domain much better. To do so, we propose and investigate three different approaches: triplet loss (contrastive finetuning) \cite{10.1007/978-3-319-24261-3_7}, Barlow Twins loss (non-contrastive finetuning) \cite{zbontar2021barlow}, and a combination of triplet loss and Barlow Twins loss (combined finetuning). To the best of our knowledge, neither the Barlow Twins loss, nor a combination of contrastive and non-contrastive losses have been applied to this task before.


\subsection{Contrastive Approach: Triplet Network}

\begin{figure}[h]
    \centering
    \includegraphics[width=\linewidth]{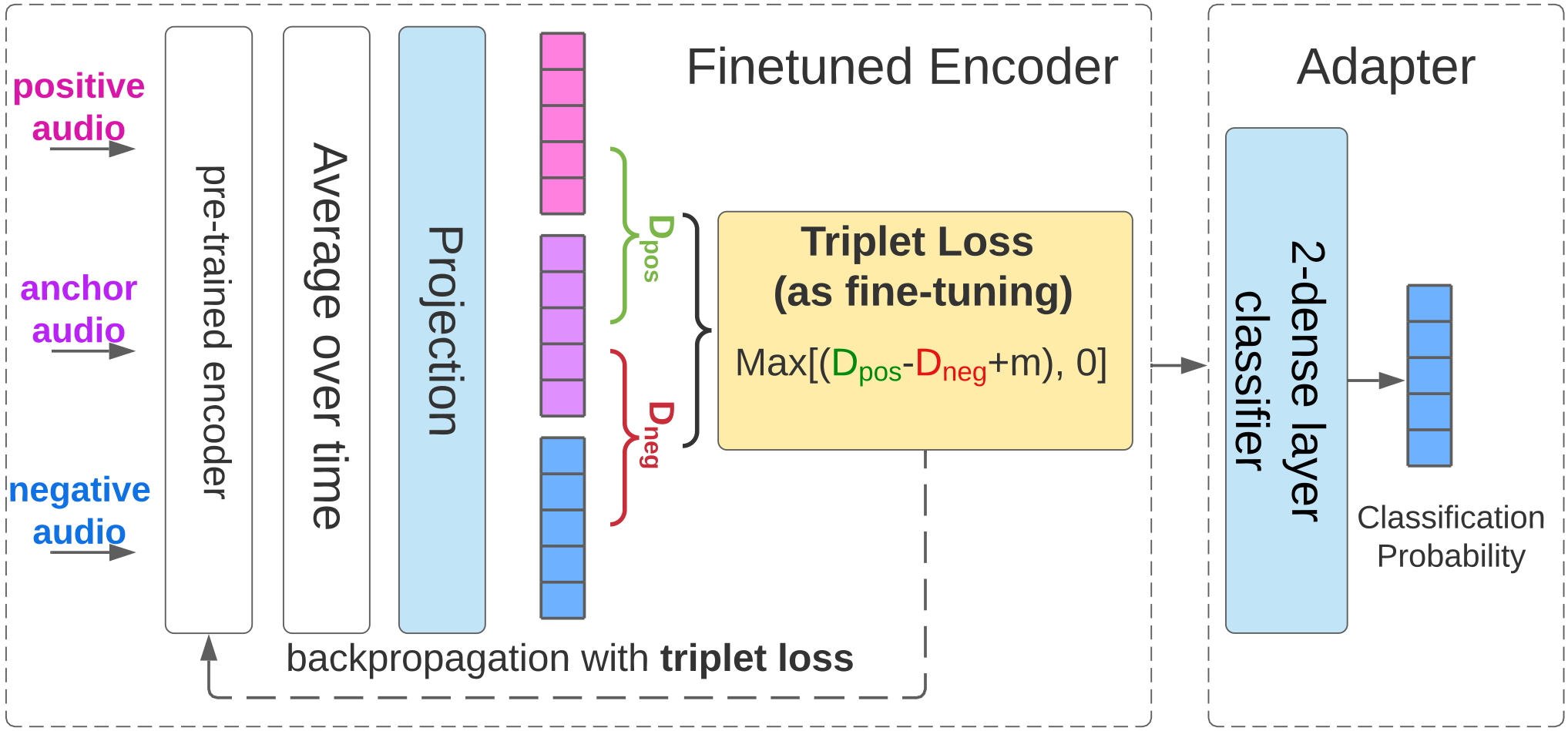} 
    \caption{Flowchart of the proposed system using contrastive finetuning.}
    \label{fig:contrasitive}
\end{figure}

Figure \ref{fig:contrasitive} shows the triplet loss \cite{10.1007/978-3-319-24261-3_7} as finetuning approach for the encoder. The triplet loss makes use of an anchor sample that will be compared to a positive sample (same class) and a negative sample (different class). Assuming encoder outputs of anchor ($e^a$), positive ($e^p$) and negative ($e^n$), the distance $D_{pos} = \lVert e^a_b - e^p_b\rVert^2$ between the anchor and positive is minimized, and the distance $D_{neg} = \lVert e^a_b - e^n_b\rVert^2 $ between anchor and negative is maximized. The triplet loss $L_{triplet}$ is illustrated in Equation \ref{eq:1}. $b$ indexes the batch and $m$ is a task specific margin as the minimum offset between distances of similar vs dissimilar pairs. 


\begin{equation}
\begin{aligned}
\label{eq:1}
L_{triplet}(e^a, e^p, e^n) = \sum_{b} max( D_{pos} - D_{neg} + m, 0 )\\ 
\end{aligned}
\end{equation}

\subsection{Non-Contrastive Approach: Barlow Twins} 
\begin{figure}[h]
    \centering
    \includegraphics[width=\linewidth]{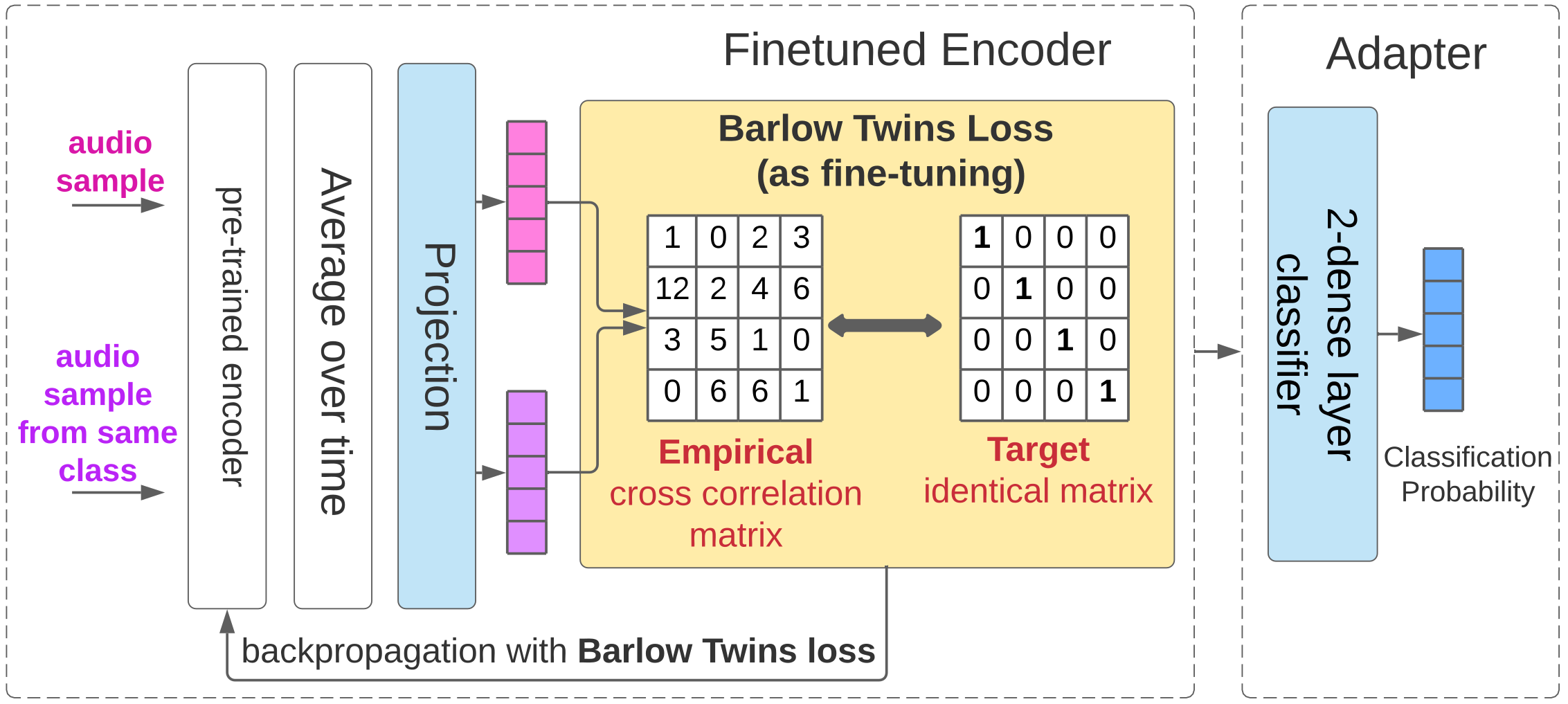} 
    \caption{Flowchart of the proposed non-contrastive learning system (Barlow Twins \cite{zbontar2021barlow}).}
    \label{fig:BarlowTwins}
\end{figure}

Figure \ref{fig:BarlowTwins} illustrates the application of the Barlow Twins approach in our proposed system. In the original implementation, the authors paired the distortion of an image sample and the image sample itself as input. However, we believe that the distortion of an audio could impact the class w.r.t emotion, age, and gender of the audio we want to embed, since voice quality and energy distribution over the pitch range are closely linked to those properties. Thus, in order to adapt Barlow Twins to the speech domain, we pair different samples from the same class as the non-contrastive input.

Formally, we calculate the cross-correlation score $C$ normalized over a batch $b$ between two vectors x and y as shown in Equation \ref{eq:2}. $x$ and $y$ correspond to pairwise encoder outputs $e^a$ and $e^p$ that are indexed with $i$ and $j$, that stem from the anchor and positive sample respectively in Equation \ref{eq:3}. 

\begin{equation}
\begin{aligned}
\label{eq:2}
C(x, y) = \frac{\langle x, y\rangle_b}{\lVert x \rVert _2\cdot \lVert y\rVert _2}
\end{aligned}
\end{equation}

Equation \ref{eq:3} defines the non-contrastive loss between the anchor encoding and the positive encoding as defined above, which corresponds to the deviation of the cross correlation matrix of the two different embeddings that have the same class label from the identity matrix. The calculation is split into a term for the main diagonal and one for the off-diagonal. Intuitively, we try to equate the diagonal elements of the cross-correlation matrix to 1 (i.e., perfect correlation) while keeping all off-diagonal values at 0 (i.e., perfect anti-correlation). This decorrelates all dimensions of the embedding, making sure that while every dimension is informative, there is no redundancy between them.

\begin{equation}
\begin{aligned}
\label{eq:3}
L_{bt}(e^a,e^p) = &\sum\limits_{i} (1 - C(e^a_i, e^p_i))^2 \\
                &+ \lambda\sum\limits_{i} \sum\limits_{j} \begin{cases} C(e^a_i,e^p_j)^2 & j\neq i \\ 0 & \text{else}\end{cases}
\end{aligned}
\end{equation}

\subsection{Combined Approach: Triplet Network \& Barlow Twins} 

\begin{equation}
\label{eq:4}
L_{com}(e^a, e^p, e^n) = \cdot L_{triplet}(e^a, e^p, e^n) + \beta \cdot L_{bt}(e^a, e^p)
\end{equation}

The combination of the two approaches introduced previously is very straightforward. We sample a triplet of positive, negative and anchor. Then we encode each of the datapoints and then feed the full triplet to the contrastive loss. The anchor and positive embeddings are fed to the non-contrastive loss. The losses are summed up with a weighting factor $\beta$ to scale $L_{bt}$ so that they are in the same order of magnitude. This is shown in Equation \ref{eq:4}. We choose $\beta = 0.01$.

\section{Experiments} \label{sec:experiments}

\subsection{Data Used}
In our experiments on SER, we use the Interactive Emotional Motion Capture (IEMOCAP) \cite{IEMOCAP2008} dataset for both finetuning and evaluation with the same splits as the SUPERB benchmark \cite{yang21c_interspeech}. For age and gender classification we use a randomly chosen subset of CommonVoice \cite{commonvoice:2020} as finetuning data. We evaluate these two tasks on TIMIT \cite{timit} in order to be able to compare with related work. We do not include any data from TIMIT in our finetuning data. For the SID task, we use LibriSpeech \cite{7178964}, since unfortunately the audio portion of the VoxCeleb dataset \cite{nagrani2017voxceleb}, which is a more commonly used benchmark for the SID task, is unavailable.

\subsection{Setup}
As baseline for each task, we will finetune the wav2vec 2.0 pretrained model as well as the HuBERT pretrained model with one-stage end-to-end finetuning like in most other works. For quantitative experiments, we evaluate our speech embedding on SER, age classification, gender classification, and SID. For qualitative experiments and further analysis we calculate cluster integrity metrics and visually inspect the embedding space using t-SNE visualization \cite{Maaten2008VisualizingDU}.

\subsubsection{Speech Emotion Classification}
To compare our proposed approach with other papers, we finetune both of our backbone models on IEMOCAP using the exact splits used in the SUPERB benchmark \cite{yang21c_interspeech}, in order to ensure comparability. We evaluate accuracy (ACC) on four emotion classes: anger, neutral, happiness and sadness. Like the selection of the training data, the evaluation also follows the SUPERB benchmark precisely. We use $m=1$ to calculate the triplet loss.

\subsubsection{Age and Gender Classification}
We use the twenties, thirties, forties, fifties, and above sixties as labels for age classification. For gender classification we use male and female as labels, since that is how the datasets we are employing are annotated. The age classification task can bring interesting insights, because the task has particularly fuzzy classes which overlap each other significantly due to the rather arbitrary boundaries that do not correspond with a significant shift in the datapoints. While the classes of gender identification from speech are much more well defined, since the datasets used for training and evaluation only contain speakers with a binary gender label, the task may expose potential irregularities in the latent space with it being a binary classification task. As the training set for both of those tasks, we randomly select 30\% of the CommonVoice corpus with the amount of samples per class being balanced. We compare our three proposed approaches with our end-to-end baseline as well as related work using x-vector and d-vector \cite{s21144785}. They did however train on the full CommonVoice dataset, as well as VoxCeleb and TIMIT, so our aim is not to outperform, but to perform comparably while being much more data efficient, which is one of the underlying benefits of our proposed approach. The evaluation metric for gender is accuracy and the evaluation metric for age is the mean absolute error (MAE). Here we use a margin of $m=1$ for gender and $m=1.2$ for age in the contrastive loss.

\subsubsection{Speaker Identification}
We experiment with SID on the clean-100 subset of LibriSpeech by splitting the utterances of 247 speakers into train and test set, since we chose to go for closed-set identification. Zero-shot identification under the learn to compare framework would also be possible, as mentioned earlier. What is interesting about this task is the extremely high amount of classes compared to the other tasks. The evaluation metric is accuracy. The hyperparameter $m$ is set to $1$ in the contrastive loss.

\section{Results}

\subsection{Quantitative Results and Analysis}

\subsubsection{Speech Emotion Classification}

\begin{table}[h]
    \centering
    \begin{tabular}{@{}lcc@{}}
        \toprule
        Model - wav2vec 2.0 based & \begin{tabular}[c]{@{}c@{}}{IEMOCAP}\\ Accuracy\end{tabular} \\ \midrule
        Contrastive finetuning              & 63.64\% \\ 
        Non-Contrastive finetuning & 52.07\% \\ 
        Combined finetuning & \textbf{67.17}\%                                                    \\ \midrule
        naive end-to-end finetuning           & 58.34\% \\
        wav2vec 2.0 Base \cite{yang21c_interspeech}     & 63.43\% \\
        wav2vec 2.0 Large \cite{yang21c_interspeech}    & 65.64\% \\ \bottomrule
    \end{tabular}
        \caption{Speech Emotion Classification on the {SUPERB} benchmark \cite{yang21c_interspeech} compared to the best wav2vec 2.0 based models in the challenge. We use the Base variant of wav2vec 2.0 as our backbone.}
    \label{table:ESC_w}
\end{table}

As shown in Table \ref{table:ESC_w}, the combined approach is better than the wav2vec 2.0 classification result from the {SUPERB} leaderboards, even in the case of the Large variant, whereas we use the Base variant as encoder. Through using the same encoder and an even simpler classifier on top of it than was used in the related work, we show that our two-step finetuning method is well suited for adjusting a general purpose speech embedding to a task specific one. Moreover, the combined approach achieves a higher score than the contrastive or non-contrastive ones, which supports our hypothesis that combining contrastive and non-contrastive approaches can remedy each other's shortcomings. The naively end-to-end finetuned model also shows further potential for our approaches if tuned more carefully, since there is a large gap between this baseline and the baseline we took from the SUPERB leaderboards.

\begin{table}[h]
    \centering
    \begin{tabular}{@{}lcc@{}}
        \toprule
        Model - HuBERT based & \begin{tabular}[c]{@{}c@{}}{IEMOCAP}\\ Accuracy\end{tabular} \\ \midrule
        Contrastive finetuning              & 64.80\% \\ 
        Non-Contrastive finetuning & 53.17\% \\ 
        Combined finetuning & \textbf{67.01}\%                                                    \\ \midrule
        naive end-to-end finetuning           & 59.64\% \\
        HuBERT Base \cite{yang21c_interspeech}     & 64.92\% \\
        HuBERT Large \cite{yang21c_interspeech}    & 67.62\% \\ \bottomrule
    \end{tabular}
        \caption{Speech Emotion Classification on the {SUPERB} benchmark \cite{yang21c_interspeech} compared to the best HuBERT based models in the challenge. We use the Base variant of HuBERT as our backbone.}
    \label{table:ESC_h}
\end{table}

When it comes to the HuBERT based model, we see very similar trends in Table \ref{table:ESC_h}. While we do not outperform the Large variant with the Base variant as encoder, we reduce the gap by 2.09\% in terms of absolute accuracy. Again, the combined finetuning approach performs the best out of all of the training methods and even outperforms the score reported for the Base variant in the SUPERB leaderboards.

\subsubsection{Age and Gender Classification}

\begin{table}[h]
    \centering
    \begin{tabular}{@{}lcc@{}}
    \toprule
    Model  & \begin{tabular}[c]{@{}c@{}}Gender\\ Accuracy\end{tabular} & \begin{tabular}[c]{@{}c@{}}Age\\ MAE\end{tabular} \\ \midrule
    \multicolumn{3}{c}{wav2vec 2.0 encoder} \\ \midrule
    Contrastive finetuning                 & 96.92\%   & 5.58  \\ 
    Non-Contrastive finetuning  & 97.55\%  & 6.40   \\ 
    Combined finetuning  & \textbf{98.53\%}  & \textbf{5.23}   \\ 
    naive end-to-end finetuning & 96.84\%  & 5.66 \\ \midrule
    \multicolumn{3}{c}{HuBERT encoder} \\ \midrule
    Contrastive finetuning                 & 97.26\%   & 5.39  \\ 
    Non-Contrastive finetuning  & 97.91\%  & 6.29   \\ 
    Combined finetuning  & \textbf{99.64\%}  & \textbf{5.19}   \\ 
    naive end-to-end finetuning & 97.77\%  & 5.52 \\ \midrule 
    \multicolumn{3}{c}{Related Works} \\ \midrule
    d-Vector \cite{s21144785} & 99.60\%  & 5.54   \\
    x-Vector \cite{s21144785} & 99.60\%  & 5.21   \\\bottomrule
    \end{tabular}
         \caption{Age and gender classification: Trained on 30\% data of CommonVoice and evaluated on TIMIT. Related works have trained on hundreds of hours of additional data, including data from the TIMIT corpus. The evaluation data is the same.}
    \label{table:GenderAgeClassification}
\end{table} 

Again, we can see that the combination method shows the best scores among the proposed approaches shown in Table \ref{table:GenderAgeClassification}. Interestingly, the non-contrastive approach performs fairly well on the binary task considering its lower performance on all other tasks when compared to the contrastive approach. Another interesting observation is that the exact same trend can be seen for both the wav2vec 2.0 based model, as well as the HuBERT based model. This supports our hypothesis that the approach we propose is agnostic to the backbone that is used and can bring further improvements whenever a better universal encoder model is proposed in the future.

The related works on d-Vector and x-Vector \cite{s21144785} are designed specifically to do age and gender classification in a highly tuned way and have seen a lot more task specific training data. These works are pretrained on VoxCeleb1 and finetuned on CommonVoice-train and TIMIT-train to get the best results. Still, we are able to achieve comparable results with our wav2vec 2.0 based model, performing only $1.07\%$ worse for gender accuracy and $0.02\%$ worse for age MAE. When using the HuBERT based model, we are even outperforming the related works, which underlines how our approach can use finetuning-data in a highly efficient way.

\subsubsection{Speaker Identification}

\begin{table}[h]
    \centering
    \begin{tabular}{@{}lc@{}}
    \toprule
    Model & \begin{tabular}[c]{@{}c@{}}Speaker Identification\\ Accuracy\end{tabular} \\ \midrule
    \multicolumn{2}{c}{wav2vec 2.0 encoder} \\ \midrule
   Contrastive finetuning   & 84.32\%                                                                  \\ 
    Non-Contrastive finetuning                     & 64.21\%                                                                  \\
   Combined finetuning & \textbf{85.09\%}                                                                  \\ 
    naive end-to-end finetuning  & 75.49\% \\ \midrule
    \multicolumn{2}{c}{HuBERT encoder} \\ \midrule
   Contrastive finetuning   & 84.84\%                                                                  \\ 
    Non-Contrastive finetuning                 & 66.56\%                                                                  \\
   Combined finetuning & \textbf{87.12\%}                                                                  \\ 
    naive end-to-end finetuning  & 76.73\% \\\bottomrule
    \end{tabular}
        \caption{SID: Trained and evaluated on the same speakers from LibriSpeech in a closed-set scenario.}
    \label{table:SID}
\end{table}

\begin{figure*}[!h]
  \centering
    \begin{tabular}{ccc}
        \toprule
        Contrastive Pretraining & Non-Contrastive Pretraining & Combined Pretraining \\
        \midrule \includegraphics[width = .3\textwidth]{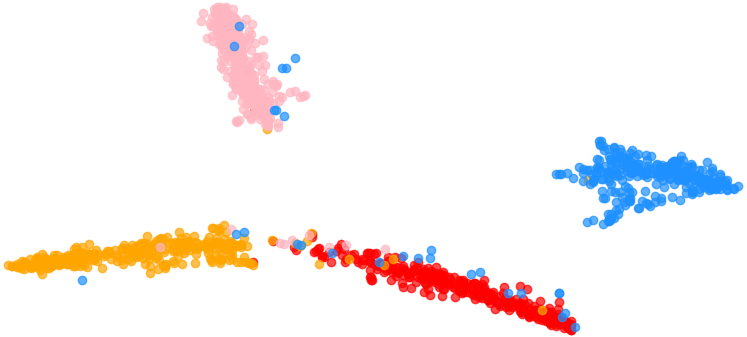} \hfill & \includegraphics[width = .3\textwidth]{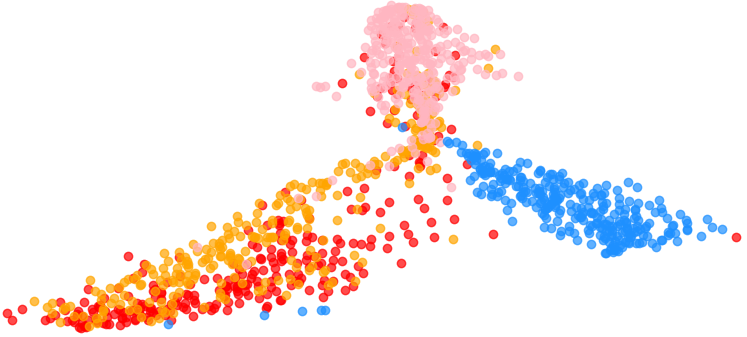} \hfill & \includegraphics[width = .3\textwidth]{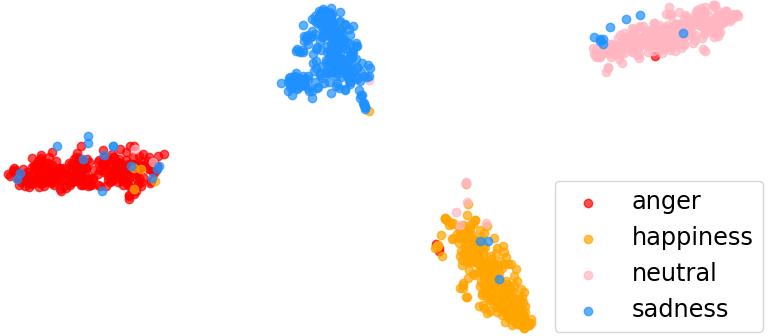}\\
        \bottomrule
    \end{tabular}
        \caption{t-SNE visualization of the emotion speech embedding latent space of a wav2vec 2.0 backbone finetuned on IEMOCAP with each method respectively. 300 samples per class from IEMOCAP are shown.}
        \label{fig:vis}
\end{figure*}

Table \ref{table:SID}  shows that even for very high amounts of classes the proposed combined approach outperforms the contrastive and non-contrastive approaches, as well as the end-to-end baseline, regardless of the backbone used. Despite the non-contrastive approach exhibiting very bad performance on this task, which is likely due to the high amount of classes, it does still improve the performance when combined with the contrastive approach. Also, we can once more observe that the HuBERT backbone leads to slightly higher scores overall that the wav2vec 2.0 backbone. Another interesting fact that can be seen from these results is that the naive finetuning baseline is outperformed in either case by almost 10\%, which is a very large margin. When visually inspecting the clusters in the latent space that each approach produces, we notice that in the latent space of the naive finetuning method, the clusters of multiple classes which correspond to similar sounding speakers are often completely collapsed. This is mostly remedied by the contrastive loss term in the finetuning process, which is then enhanced further with the non-contrastive loss promoting class invariance. So the more classes we have in a task, the more beneficial our proposed finetuning method seems to be.

\subsection{Qualitative Results and Analysis}

\begin{table}[h]
    \centering
    \begin{tabular}{@{}lcc@{}}
    \toprule
    \multirow{2}{*}{Approach} & \multicolumn{2}{c}{Distance Metrics} \\ \cmidrule(l){2-3} 
                              & Invariant Dist.  & Davies-Bouldin Index  \\ \midrule
    \multicolumn{2}{c}{wav2vec 2.0 encoder}& \\ \midrule
    Contrastive              & 2.459    & 3.086  \\ 
    Non-Contrastive          & 1.545    & 3.778  \\ 
    Combined               & \textbf{0.148}   &  \textbf{1.812}    \\ \midrule
    \multicolumn{2}{c}{HuBERT encoder}& \\ \midrule
    Contrastive              & 2.291    & 2.875  \\ 
    Non-Contrastive          & 1.388    & 3.638  \\ 
    Combined               & \textbf{0.137}   &  \textbf{1.419}    \\ \bottomrule
    \end{tabular}
        \caption{Distance metrics for IEMOCAP classes: Invariant distance measures the average euclidean distance to the centroid within a class (lower invariant distance $\rightarrow$ higher class invariance). The Davies-Bouldin index measures the variance between classes in context: clusters which are farther apart and less dispersed will result in a lower score.}
    \label{table:Averaged centroid distance}
\end{table}

The t-SNE visualization \cite{van2008visualizing} in Figure \ref{fig:vis} shows that the contrastive approach is able to form distinct clusters in the embedding space, which are however spread fairly wide and almost overlap at the border between happy and angry. The non-contrastive approach shows tendencies of working as intended with clusters clearly visible, however they overlap greatly. It seems like the Barlow Twins approach is not robust against the mode-collapse problem in this configuration. The combined approach however produces distinct clusters which are more concise than the contrastive clusters, showing how the explicit variance reduction objective of Barlow Twins enriches this approach. This has strong implications on how suitable the tuned model would be in a zero-shot application, which essentially assigns a label by clustering an unseen datapoint in the latent space. For this, the more concise clusters with less overlap are intuitively better suited.

Finally, the distance evaluation in Table \ref{table:Averaged centroid distance} further validates the performance bottlenecks of the individual models and how the combined approach can solve them. The variance on the emotion classification task is much lower for the combined approach, showing that it is very well suited to be further processed by a classifier that goes on top of it, or to be directly used for zero-shot inference under the aforementioned learn-to-compare framework \cite{sung2018learning}.

\section{Conclusions}

In this paper we propose a data efficient way of improving the ability of a pretrained model for general purpose speech embeddings to be finetuned to a new task or domain. Instead of adding a classifier on top of such an encoder and finetuning it end-to-end, we propose to firstly finetune the encoder alone using a combination of contrastive and non-contrastive losses and then train an adapter to go from the adjusted representations to the final output. In our experiments we show the benefit of the two-step training over naive end-to-end finetuning, as well as the benefit of combining contrastive and non-contrastive methods for this two-step finetuning, which results in more reliable latent class representations with reduced class variance and increased discriminativeness between classes. Our approach performs competitively in various tasks which vary in their amount of classes from just 2 to over 200 and in the fuzziness of the underlying concepts of these classes. Furthermore, the single-step finetuning baseline is outperformed in every case.

While we apply this framework to the analysis of speech properties with both a wav2vec 2.0 and a HuBERT encoder, we believe that other encoders could as well benefit from our approach, even in other domains or modalities that leverage encoders which are trained in a self-supervised fashion to produce general purpose embeddings, such as word embeddings in a language model.

\newpage
\bibliographystyle{IEEEbib}
\bibliography{main}

\end{document}